\documentclass[journal]{IEEEtran}
\usepackage{amsmath,amsfonts}
\usepackage{algorithmic}
\usepackage{algorithm}
\usepackage{array}
\usepackage{textcomp}
\usepackage{stfloats}
\usepackage{url}
\usepackage{hyperref}
\usepackage{verbatim}
\usepackage{graphicx}
\usepackage[nocompress]{cite}
\usepackage{subfigure}
\usepackage{multirow}
\usepackage{epsfig}
\usepackage{amsmath}
\usepackage{amssymb}
\usepackage{wrapfig}
\usepackage{booktabs}
\usepackage{pifont}
\usepackage{float}
\usepackage{abbrv}
\usepackage[table]{xcolor}
\usepackage{colortbl}
\usepackage{multicol}
\usepackage{multirow}
\usepackage{makecell}
\definecolor{lightgreen}{RGB}{232,255,232}

\hypersetup{
  colorlinks=true,
  linkcolor=red,
  citecolor=blue,
}
\begin{document}

\title{ChangeViT: Unleashing Plain Vision Transformers for Change Detection}
\author{Duowang~Zhu$^*$, Xiaohu~Huang$^*$, Haiyan~Huang, Zhenfeng~Shao$^\dagger$, and Qimin~Cheng
\thanks{Duowang~Zhu, Haiyan~Huang and Zhenfeng~Shao are with the State Key Laboratory of Information Engineering in Surveying, Mapping and Remote Sensing, Wuhan University, Wuhan 430079, China (email: zhuduowang@whu.edu.cn; huanghaiyan@whu.edu.cn; shaozhenfeng@whu.edu.cn).}
\thanks{Xiaohu~Huang is with the University of Hong Kong, Pokfulam, Hong Kong (e-mail: huangxiaohu@connect.hku.hk)}
\thanks{Qimin~Cheng is with the School of Electronic Information and Communications, Huazhong University of Science and Technology, Wuhan 430074, China (e-mail: chengqm@hust.edu.cn).}
\thanks{$^*$: Equal contribution.}
\thanks{$^\dagger$: Corresponding author.}
}

\markboth{JOURNAL OF \LaTeX\ CLASS FILES, VOL. XX, NO. XX}
{Zhu \MakeLowercase{\textit{et al.}}: ChangeViT: Unleashing Plain Vision Transformers for Change Detection}

\maketitle

\begin{abstract}
Change detection in remote sensing images is essential for tracking environmental changes on the Earth’s surface. Despite the success of vision transformers (ViTs) as backbones in numerous computer vision applications, they remain underutilized in change detection, where convolutional neural networks (CNNs) continue to dominate due to their powerful feature extraction capabilities. In this paper, our study uncovers ViTs’ unique advantage in discerning large-scale changes, a capability where CNNs fall short. Capitalizing on this insight, we introduce ChangeViT, a framework that adopts a plain ViT backbone to enhance the performance of large-scale changes. This framework is supplemented by a detail-capture module that generates detailed spatial features and a feature injector that efficiently integrates fine-grained spatial information into high-level semantic learning. The feature integration ensures that ChangeViT excels in both detecting large-scale changes and capturing fine-grained details, providing comprehensive change detection across diverse scales. Without bells and whistles, ChangeViT achieves state-of-the-art performance on three popular high-resolution datasets (\ie, LEVIR-CD, WHU-CD, and CLCD) and one low-resolution dataset (\ie, OSCD), which underscores the unleashed potential of plain ViTs for change detection. Furthermore, thorough quantitative and qualitative analyses validate the efficacy of the introduced modules, solidifying the effectiveness of our approach. The source code is available at \url{https://github.com/zhuduowang/ChangeViT}.
\end{abstract}


\section{Introduction}
\label{sec:introduction}
\IEEEPARstart{C}{hange} detection plays a crucial role in the field of remote sensing, employing pairs of bi-temporal images taken of the same geographic area at different times to track changes on the Earth’s surface over time \cite{radke2005image}. It has been widely applied in various applications such as disaster assessment \cite{zheng2021building}, urban planning \cite{wang2021land}, arable land protection \cite{lunetta2022land}, and environmental management \cite{kennedy2009remote}. In recent years, convolutional neural networks (CNNs) have emerged as the primary backbone choice for state-of-the-art change detectors \cite{zheng2021building, liu2023attention, zhang2023global, feng2023lightweight, li2023lightweight, zheng2021change}, as they can extract rich hierarchical features for detecting changes with different sizes.

Over the past few years,  Vision Transformers (ViTs) \cite{dosovitskiy2020image} have \textit{de facto} substituted CNNs as the dominant backbones in various computer vision tasks, \eg, object detection \cite{li2022exploring}, image segmentation \cite{liu2023simpleclick}, image matting \cite{yao2024vitmatte}, and pose estimation \cite{xu2022vitpose}, which exhibit superior performance than CNN-based methods benefiting from the long-range modeling capability. While transformers have been explored in the context of change detection in some preliminary studies \cite{liu2023attention, chen2021remote, jiang2023vct, bandara2022transformer, zhao2023adapting_AdapterCD}, their performance has not yet matched that of the leading CNN models. Therefore, this paper aims to study the potential benefits of ViTs for change detection, striving to unleash their effectiveness in this area.

\begin{figure}[t]
    \centering
    \subfigure[Performance comparison of different change detectors.]{
    \centering
    \includegraphics[width=0.8\linewidth]{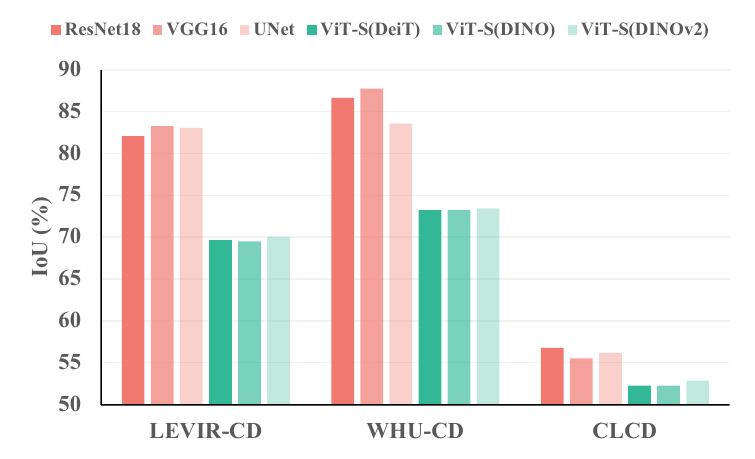}
    \label{subfig:investigate_different_methods_v4}
    }
    \centering
    \subfigure[Performance comparison between CNN and ViT models for various change sizes.]{
    \centering
    \includegraphics[width=0.96\linewidth]{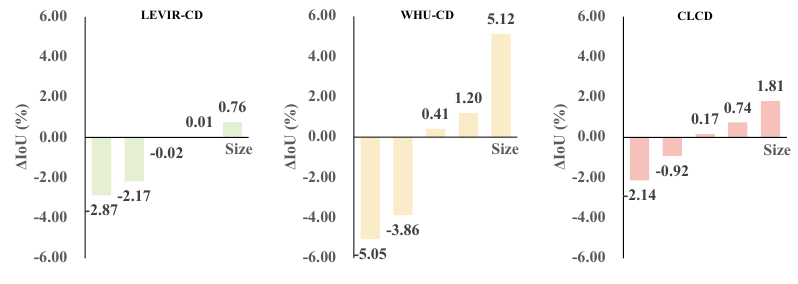}
    \label{subfig:intro_cds_v2c}
    }
    \caption{(a) Performance comparison of different change detectors across three datasets, categorized as CNN-based and ViT-based models. (b) Performance comparison ($\Delta$IoU (\%)) between a CNN (ResNet18) and a ViT (ViT-S (DINOv2)) model for detecting changes with various sizes. The horizontal axis incrementally reflects the change sizes, progressing from smallest to largest changes. The $\Delta$IoU values presented are calculated by subtracting the CNN’s performance from that of the ViT for each size category.}
    \label{fig:motivation}
\end{figure}

To assess the efficacy of ViTs in the change detection task, we first conduct a comprehensive performance comparison between change detectors utilizing ViTs and three established CNN architectures as backbones, \ie, ResNet18 \cite{he2016deep}, VGG16 \cite{simonyan2014very}, and UNet \cite{ronneberger2015u_unet}. This evaluation spans three well-known datasets, \ie, LEVIR-CD \cite{chen2020spatial}, WHU-CD \cite{ji2018fully}, and CLCD \cite{liu2022cnn_clcd}, as depicted in Fig.~\ref{subfig:investigate_different_methods_v4}. Additionally, we explore the influence of various model initializations by incorporating pre-trained weights from DeiT \cite{touvron2021training_deit}, DINO \cite{dino}, and DINOv2 \cite{oquab2023dinov2} into our analysis. Specifically, ResNet18, VGG16, UNet, and ViT-S (DeiT) are pre-trained on ImageNet-1k with supervised training, while ViT-S(DINO) and ViT-S(DINOv2) are pre-trained with self-supervised training on ImageNet-1k, ImageNet-22k, and Google Landmarks, \etc. The results indicate that: (1) CNN models significantly outperform all ViTs across all datasets, regardless of whether supervised or self-supervised learning is used, highlighting the dominance of CNNs in change detection tasks. (2) Even with identical data initialization (\ie, ImageNet-1k), the performance of ViTs remains inferior to that of CNN-based models.

To delve deeper into the models’ capabilities, we perform an in-depth analysis of a ViT (ViT-S with DeiT pre-training) and a CNN model (ResNet18 with ImageNet-1k pre-training) in detecting changes across various object sizes, which is illustrated in Fig.~\ref{subfig:intro_cds_v2c}. We organize the test samples from each dataset by the proportion of pixels occupied by different objects within the images. Specifically, we first sort the images in ascending order based on the ratio of pixels occupied by changing objects to the total number of pixels in the image. Then, we evenly divide this ordered sequence into five categories, ranging from the smallest to the largest proportions. We calculate the average performance difference between the ViT and CNN models within each category. The results show that though ViTs lag behind CNNs in detecting smaller changes, they demonstrate enhanced reliability for larger objects across all datasets. These insights suggest that while ViTs cannot capture fine-grained details as effectively as CNNs, they excel in detecting large-scale changes. Therefore, this previously untapped benefit has the potential to effectively mitigate the limitations inherent in CNN architectures.

Building upon the insights gathered from our preceding analysis, we propose \textbf{ChangeViT}, a simple yet effective framework that leverages the plain ViT framework as its core to capture large-scale object information. This is coupled with a detail-capture module specifically used to focus on fine-grained features. The detail-capture module functions as an auxiliary network, incorporating selected layers (C2-C4) from ResNet18 \cite{he2016deep}, which offers a more compact footprint (2.7M parameters) compared to a complete CNN model (11.2M parameters). To seamlessly inject these fine-grained details into the feature representation of ViTs, we establish connections between ViT’s representations and fine-grained features. This integration is accomplished by considering ViT features as queries and merging the fine-grained features by applying the cross-attention mechanism.

Through extensive experiments on four widely recognized datasets, \ie, LEVIR-CD \cite{chen2020spatial}, WHU-CD \cite{ji2018fully}, CLCD \cite{liu2022cnn_clcd}, and OSCD \cite{daudt2018urban_OSCD}, ChangeViT achieves the state-of-the-art performance across the board. In addition, we combine the proposed modules with various hierarchical transformers, \ie, Swin Transformer \cite{liu2021swin}, PVT \cite{wang2021pyramid_pvt}, and PiT \cite{heo2021rethinking_pit}. Consistently across these architectures, the proposed modules enhance performance, thereby further confirming their efficacy. Notably, despite the plain ViT’s perceived limitations compared to these advanced hierarchical networks, ChangeViT outperforms methods that utilize these complex models, showcasing that we effectively unleash the capacity of plain ViTs in the field of change detection.

The main contributions of this paper can be summarized as follows:
\begin{itemize}
\item We thoroughly investigate the performance of plain ViTs and identify their aptitude for detecting large-scale changes. Motivated by this finding, we introduce ChangeViT, a simple yet effective framework which utilizes plain ViT as the primary feature extractor for the change detection task.

\item To enhance the detection of changes across various sizes, we integrate a detail-capture module, specifically introduced to address the limitations of ViTs when identifying small objects. Furthermore, we introduce a feature injector to merge the extracted detailed features into high-level ones from the ViT, ensuring comprehensive feature representation within the model.

\item ChangeViT achieves state-of-the-art performance on four popular datasets, \ie, LEVIR-CD, WHU-CD, CLCD, and OSCD, demonstrating the superiority of the proposed method. Moreover, thorough quantitative and qualitative analyses validate the efficacy of the modules we have introduced, further solidifying the effectiveness of our approach.

\end{itemize}

\section{Related Work}
\label{sec:related_work}

\subsection{Change Detection}
Regarding the network architecture, existing change detection methods employing deep learning can be generally categorized into two groups: CNN-based and transformer-based.

\textbf{CNN-based Methods.} CNN-based change detection approaches have been the mainstream framework in the literature \cite{zhang2023global, li2023lightweight, feng2022icif, fang2023changer, daudt2018fully, feng2023change, feng2023lightweight, zhang2023aernet, liu2020building, daudt2018urban} for a long time, known for their hierarchical feature modeling capabilities. These works primarily focus on multi-scale feature extraction, difference modeling, lightweight architecture designing, and foreground-background class imbalance. For instance, methods in \cite{daudt2018fully, daudt2018urban} utilize fully convolutional networks to capture hierarchical features for learning multi-scale feature representations. For adequate differential feature modeling, approaches in \cite{feng2022icif, feng2023change} incorporate the attention mechanism to establish relational dependencies among bi-temporal features. In contrast, Changer \cite{fang2023changer} introduces a parameter-free method, which simply exchanges the characteristics of each phase to capture and perceive each other's information. Methods in \cite{li2023lightweight, feng2023lightweight} focus on designing efficient and effective network architectures, utilizing lightweight feature extractors \cite{sandler2018mobilenetv2, tang2022ghostnetv2} as backbones. Several studies \cite{zhang2023global, zhang2023aernet} address the significant challenge posed by foreground-background class imbalance by developing innovative loss functions that prioritize foreground alterations while minimizing interference from background noise (\eg, seasonal variations, climate changes).

\textbf{Transformer-based Methods.} Recently, Vision Transformer \cite{dosovitskiy2020image} and its variants \cite{liu2021swin, liu2022swin, wang2021pyramid_pvt} have surpassed CNN in various visual tasks and became the dominant backbone \cite{li2022exploring, yao2024vitmatte, xu2022vitpose, chen2023vision}. Motivated by these achievements, several works \cite{chen2021remote, bandara2022transformer, liu2023attention, ma2024eatder, zhang2022swinsunet, jiang2023vct, zhao2023adapting_AdapterCD} have explored the application of transformers in change detection tasks. Some of these methods \cite{zhang2022swinsunet, bandara2022transformer} utilize pure transformers, while others \cite{chen2021remote, liu2023attention, ma2024eatder, jiang2023vct} adopt CNN-Transformer hybrid architectures. Methods in  \cite{zhang2022swinsunet, bandara2022transformer} introduce hierarchical transformer networks based on the swin transformer \cite{liu2021swin}. The others typically follow a paradigm in which features extracted by CNN serve as semantic tokens, followed by contextual relation modeling between bi-temporal tokens using transformer blocks. The method introduced in \cite{zhao2023adapting_AdapterCD} exhibits an efficient tuning strategy that involves freezing the parameters of the Transformer encoder while introducing additional trainable parameters. However, this method fails to deliver optimal results due to an inadequate exploration of the strengths and limitations of the transformers. This precludes a more effective application of the model’s capabilities, thereby capping the potential gains in performance.

Different from the previous approaches mainly using hierarchical networks, the proposed ChangeViT applies the plain ViT as the cornerstone feature extractor, which we find has previously unidentified potential in detecting large-scale changes. 

\begin{figure*}[htb]
\centering
\includegraphics[width=0.9\linewidth]{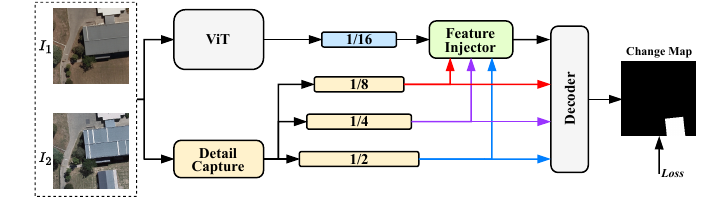}
\caption{Overview of the proposed \textbf{ChangeViT}. bi-temporal images $I_{1}$ and $I_{2}$ are firstly fed into shared ViT to extract high-level semantic features and detail-capture module to extract low-level detailed information. Subsequently, a feature injector is introduced to inject the low-level details into high-level features. Finally, a decoder is utilized to predict changed probability maps.
}
\label{fig:framework}
\end{figure*}

\subsection{Plain ViT for Downstream Tasks}
ViT \cite{dosovitskiy2020image} is a plain, non-hierarchical architecture, which is a powerful alternative to standard CNN for image classification. Due to the significant computational overhead of self-attention in ViT, subsequent works focus on designing more efficient architectures, such as Swin \cite{liu2021swin}, PVT \cite{wang2021pyramid_pvt} and PiT \cite{heo2021rethinking_pit}. These works inherit some designs from CNN, including hierarchical structures, sliding windows, and convolutions. Recently, researchers have begun to study the potential of ViT for various downstream tasks motivated by the emergence of large pre-trained models, \eg, DeiT \cite{touvron2021training_deit}, DINO \cite{caron2021emerging_dino}, DINOv2 \cite{oquab2023dinov2}, MAE \cite{he2022masked} and CLIP \cite{radford2021learning}. The plain ViT has already made remarkable progress in dense prediction \cite{chen2023vision, li2022exploring, liu2023simpleclick}, pose estimation \cite{xu2022vitpose}, image matting \cite{yao2024vitmatte}, \etc. ViTDet \cite{li2022exploring} is the first to employ plain, non-hierarchical ViT as the backbone for object detection with minimal adaptation, \ie, building a simple feature pyramid for single-scale features and aiding a few cross windows for information propagation. ViT-Adapter \cite{chen2023vision} introduces a pre-training-free adapter that injects prior knowledge to ViT without redesigning its architecture for various dense prediction tasks. Similarly, SimpleClick \cite{liu2023simpleclick} and ViTPose\cite{xu2022vitpose} apply vanilla ViT as the feature extractor to acquire single-scale features. For image matting, ViTMatte \cite{yao2024vitmatte} is the first work to unleash the potential of ViT with concise adaption.

Inspired by the above works, we aim to unleash the potential of the plain ViT model, enabling it to adapt well to change detection tasks.

\section{Proposed Method}
\label{sec:proposed_method}

The overall architecture is illustrated in Fig.~\ref{fig:framework}. For the bi-temporal images $I_{1} \in \mathbb{R}^{H \times W \times 3}$ and $I_{2} \in \mathbb{R}^{H \times W \times 3}$, they are parallelly fed into a ViT and a detail-capture module. The ViT extracts high-level features $F_{V}^{t} \in \mathbb{R}^{\frac{H}{16} \times \frac{W}{16} \times C_{4}}$, where $t \in \{1,2\}$ represents two phases, while the detail-capture module acquires fine-grained multi-scale features $F_{C_{i}}^{t} \in \mathbb{R}^{\frac{H}{2^{i}} \times \frac{W}{2^{i}} \times C_{i}}$ ($i \in \{1,2,3\}$, $t \in \{1,2\}$). To enhance the detection of intricate details within high-level features, we introduce a feature injector aimed at integrating low-level fine-grained information into $F_{V}$. Finally, a multi-scale feature fusion decoder is applied to predict the changed probability map $P \in \mathbb{R}^{H \times W \times 1}$. 

\subsection{Feature Extraction}
\label{subsec:feature_extraction}
The feature extractor is composed of a plain ViT, and a detail-capture module which is described as follows: 

\textbf{Plain ViT.} Bi-temporal images $I_{1}$, $I_{2}$ are fed into patch embedding layer, dividing them into non-overlapping $16 \times 16$ patches. These patches are then flattened and projected to $D$-dimension tokens, and the feature resolution is reduced to $1/16$ of the original images. Afterwards, position embedding is added to these tokens, which are passed through $L$ transformer layers. Each layer consists of a layer normalization (LN), a multi-head self-attention (MHSA) and a feed-forward network (FFN). The formulation of these layers is given by Eq.~\ref{transformer_layer_0} and Eq.~\ref{transformer_layer_1}: 
\begin{equation}
    \begin{aligned}
        F_{V}^{' t,i+1} = F_{V}^{t,i} + \mathrm{MHSA}(\mathrm{LN}(F_{V}^{t,i})),
    \end{aligned}
    \label{transformer_layer_0}
\end{equation}
\begin{equation}
    \begin{aligned}
        F_{V}^{t,i+1} = F_{V}^{' t,i+1} + \mathrm{FFN}(\mathrm{LN}(F_{V}^{' t,i+1})).
    \end{aligned}
    \label{transformer_layer_1}
\end{equation}
where $i$ denotes the output of the $i$th transformer layer. The final output of the ViT backbone is represented as $F_{V}^{t} \in \mathbb{R}^{\frac{H}{16} \times \frac{W}{16} \times C_{4}}$, where $C_{4}$ equals to $D$ . 

\textbf{Detail-capture.} As discussed in Sec.~\ref{sec:introduction}, ViT demonstrates proficiency in detecting large changes but exhibits reduced effectiveness with smaller ones. Addressing this challenge, we introduce a detail-capture module designed to compensate for the absence of fine-grained local cues crucial for change detection. This module comprises three residual convolutional blocks (C2-C4) adapted from ResNet18 \cite{he2016deep}. Upon processing the input images through the detail-capture module, three-scale detailed features are generated, \ie, $1/2$, $1/4$ and $1/8$, denoted as $F_{C_{i}}^{t} \in \mathbb{R}^{\frac{H}{2^{i}} \times \frac{W}{2^{i}} \times C_{i}}$ ($i \in \{1,2,3\}$).

\subsection{Feature Injector}
\label{subsec:structure_enhancement}
In the change detection task, preserving detailed spatial features is crucial as they can help detect small objects. Ensuring the effective transmission of low-level details to high-level semantic features is paramount. 

\begin{figure}[t]
    \centering
    \subfigure[Using ViT's features as query and detailed features as key and value for feature injector.]{
    \centering
    \includegraphics[width=\linewidth]{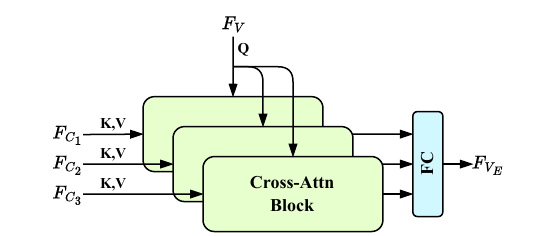}
    \label{subfig:structure_enhancement_v1}
    }
    \centering
    \subfigure[Using ViT's features as key and value and detailed features as query for feature injector.]{
    \centering
    \includegraphics[width=\linewidth]{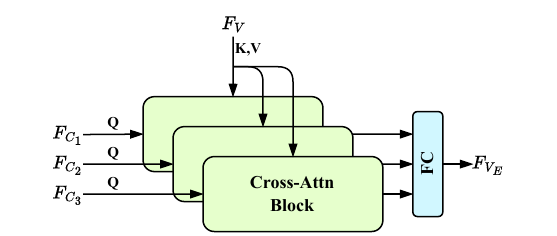}
    \label{subfig:structure_enhancement_v2}
    }
    \caption{Illustration of the feature injectors. $F_{C_i}$ ($i\in\{1,2,3\}$) denote multi-scale detailed features acquired from the detail-capture module, while $F_V$ denotes the ViT's feature lacking detailed information. (a) Let $F_V$ as the query vector, and $F_{C_i}$ as the key and value vectors to capture detailed features for ViT. (b) Using $F_V$ as the key and value vectors, and $F_{C_i}$ as the query vector to refine features for ViT.}
    \label{fig:structure_enhancement}
\end{figure}

Therefore, we introduce a feature injector, composed of three cross-attention blocks \cite{chen2021crossvit}, as illustrated in Fig.~\ref{subfig:structure_enhancement_v1}. It considers the low-level features as the key and value vectors and the high-level feature as the query vector. Intuitively, this is reasonable as it allows the feature injector to gather the most relevant information based on the provided key information and integrate it into the query. By enabling cross-layer feature propagation, detailed information can be incorporated into the high-level representations of the ViT, denoted as $F_{V_{E}}^{t}$. The $F_{V_{E}}^{t}$ is computed as follows: 
\begin{equation}
    F_{V_{E}}^{t,i} = \mathrm{CrossAttn}(F_{V}^{t}, F_{C_{i}}^{t}),
\end{equation}
\begin{equation}
    F_{V_{E}}^{t} = \mathrm{FC}(F_{V_{E}}^{t,1} \textcircled{c} F_{V_{E}}^{t,2} \textcircled{c} F_{V_{E}}^{t,3}).
\end{equation}
where $i \in \{1,2,3\}$ denotes the index of low-level layers, $F_{V}^{t}$ as query and $F_{C_{i}}^{t}$ as key and  value, respectively. The $\mathrm{FC}$ is a 2D depth-wise convolution with the kernel size of $1\times1$ and \textcircled{c} denotes concatenation operation along the channel dimension.

Additionally, we explore an alternative approach to feature injector, as depicted in Fig.~\ref{subfig:structure_enhancement_v2}, which considers the low-level features as query, and the ViT's semantic information as key and value to refine the ViT's representation according to the characteristics of the hierarchical detailed features.

\subsection{Decoder and Optimization}
\label{subsec:decoder_and_optimization}
Compared to existing methods \cite{zhang2023global, liu2023attention, feng2023change, ma2024eatder}, which employ complex techniques to model difference information and predict the change probability map, we chose a simpler decoder to better demonstrate the learning capabilities of ChangeViT. Specifically, we use a straightforward feature fusion layer to capture differences between bi-temporal features. A cascade convolutional layer, followed by an upsampling operation, is employed to progressively aggregate differential features from deep to shallow layers, ultimately restoring them to the original resolution of $H \times W$. The difference modeling is formulated as Eq.~\ref{difference_modeling}: 
\begin{equation}
    F_{D_{i}} = \mathrm{MLP}(F^{1}_{i} \textcircled{c} F^{2}_{i} \textcircled{c} \lvert{F^{1}_{i}-F^{2}_{i}}\rvert),
    \label{difference_modeling}
\end{equation}
where $F_{i}^{t} \in \{F_{C_{1}}^{t}, F_{C_{2}}^{t}, F_{C_{3}}^{t}, F_{V_{E}}^{t}\}$ ($t \in \{1,2\}$), $\mathrm{MLP}$ is a three-layer 2D convolutional network with kernel size of $3\times3$ along with ReLU activation function, \textcircled{c} denotes concatenating on channel dimension, and $|\cdot|$ means absolute value operation.

To restore the original resolution of the changed map, we use a simple cascade upsampling operation, which is represented as follows:
\begin{equation}
    F_{D_{i+1}} \leftarrow \mathrm{Deconv_{4\times4}}(\mathrm{Conv_{1\times1}}(F_{D_{i}})) + F_{D_{i+1}},
    \label{upsample}
\end{equation}
where $\mathrm{Conv_{1\times1}}$ is a 2D convolution with a kernel size of $1\times1$ to reduce the channel dimension, and $\mathrm{Deconv_{4\times4}}$ denotes 2D deconvolution to upsample the feature map with a kernel size of $4\times4$ and stride size of $2\times2$.

Finally, a classification layer is applied to transform the shallowest features $F_{D_{4}}$ into change maps $P$, which is formulated as Eq.~\ref{eq:cls_layer}:
\begin{equation}
    \centering
    P = \mathrm{Sigmoid}(\mathrm{Conv_{3\times3}}(F_{D_{4}})).
    \label{eq:cls_layer}
\end{equation}
where $\mathrm{Conv_{3\times3}}$ is a 2D convolution with the kernel size of $3 \times 3$, $\mathrm{Sigmoid}$ function maps the feature map to $(0, 1)$ and then transforms to a binary map given a predefined threshold (\ie, 0.5), \ie, $P \in \{0,1\}^{H \times W}$.

As mentioned in prior works \cite{zhang2023global, li2023lightweight}, the proportion of changed targets is significantly lower than that of unchanged ones. Following the above works, we adopt binary cross-entropy (BCE) and dice loss (Dice) \cite{milletari2016v_dice} to alleviate the class imbalance problem. The change detection loss $\mathcal{L}_{total}$ is defined as Eq.~\ref{loss_function}:
\begin{equation}
\begin{aligned}
    \mathcal{L}_{total} = \mathcal{L}_{bce}(P,Y) + \mathcal{L}_{dice}(P,Y),
    \label{loss_function}
\end{aligned}
\end{equation}
The BCE and Dice losses are formulated as follows: 


\begin{gather}
    \label{hybrid_loss}
    \mathcal{L}_{bce}(P,Y) = -\frac{1}{N}\sum_{i=1}^N[Y_i\log_2P_i + (1-Y_i)\log_2(1-P_i)], \nonumber \\
    \mathcal{L}_{dice}(P,Y) = 1 - \frac{2\sum_{i=1}^{N}{{P_i}{Y_i}}+\epsilon}{\sum_{i=1}^{N}{(P_{i})^{2}}+\sum_{i=1}^{N}{(Y_{i})^{2}}+\epsilon}.
\end{gather}
where $i$ denotes the $i$-th pixel, $N$ is the number of total pixels, $Y$ denotes the ground truth and $\epsilon$ (\ie, 1e-5) is a smooth term utilized to avoid zero division.

\section{Experiments}
\label{sec:experiments}
We conducted extensive experiments on three widely used high-resolution datasets, namely LEVIR-CD \cite{chen2020spatial}, WHU-CD \cite{ji2018fully}, and CLCD \cite{liu2022cnn_clcd}, as well as one challenging low-resolution dataset, OSCD \cite{daudt2018urban_OSCD}, to demonstrate the effectiveness of the proposed method. To better understand each component of ChangeViT, we conduct extensive diagnostic experiments in Sec.~\ref{subsec:ablation_experiments}. Otherwise stated, we use ChangeViT-S for experiments on the three high-resolution datasets.

\subsection{Implementation Details}
\label{subsec:implementation_details}
We adopt vanilla ViT \cite{dosovitskiy2020image} as our primary backbone, specifically incorporating its tiny and small variants, thereby constructing two models named ChangeViT-T and ChangeViT-S. We use DeiT \cite{touvron2021training_deit} and DINOv2 \cite{oquab2023dinov2} pre-trained weights for initialization, respectively. Our models are implemented using the PyTorch framework \cite{paszke2017automatic} and executed on a computing platform consisting of a single NVIDIA GeForce RTX 3090 GPU paired with an Intel(R) Xeon(R) Gold 6138 CPU. For optimization, we opt for the Adam optimizer \cite{kingma2014adam}, with beta values set to (0.9, 0.99) and a weight decay of 1e-4. Initially, the learning rate is 2e-4 and gradually reduces according to a scheduled reduction formula: (1-(curr\_iter/max\_iter))$^{\alpha} \times$ lr, where $\alpha$ is set to 0.9 and max\_iter is set to 80K iterations on LEVIR-CD and WHU-CD, 40K for CLCD dataset, and 10K for OSCD, respectively. The batch size remains at 16 across all experiments. To augment the training data and bolster the model's robustness, we apply random flipping and cropping data enhancement approaches. The channel dimensions of $F_{C_{i}}$ are set to 64, 128, and 256, respectively. Furthermore, we ensure consistency and fairness in comparison by meticulously aligning the experimental settings of the compared methods with those specified in the original paper.

\subsection{Datasets}
\label{subsec:datasets}
\subsubsection{LEVIR-CD}
\label{levir_cd}
This dataset \cite{chen2020spatial} comprises 637 high-resolution (1024$\times$1024, 0.5 m/pixel) bi-temporal image pairs, sourced from Google Earth. The images represent 20 diverse regions across various Texan cities, including Austin, Lakeway, Bee Cave, Buda, Kyle, Manor, Pflugervilletx, Dripping Springs, and others. The dataset, with annotations for 31333 individual building changes, spans images captured from 2002 to 2018 in various locations. Following the cropping methodology established in \cite{liu2023attention}, each image is segmented into 16 distinct 256$\times$256 patches. Consequently, the dataset is divided into 7120 pairs for training, 1024 pairs for validation, and 2048 pairs for testing.

\subsubsection{WHU-CD}
\label{whu_cd}
This publicly available dataset \cite{ji2018fully} focuses on building change detection and includes high-resolution (0.2 m) bi-temporal aerial images, totaling 32507$\times$15354 pixels. It primarily encompasses areas affected by earthquakes and subsequent reconstruction, mainly involving building renovations. Adhering to the standard procedure detailed in \cite{feng2022icif}, the dataset images are divided into 256$\times$256 non-overlapping patches. The dataset is partitioned into 5947 training pairs, 744 validation pairs, and 744 test pairs.

\subsubsection{CLCD}
\label{clcd}
The CLCD \cite{liu2022cnn_clcd} dataset consists of cropland change samples, including buildings, roads, lakes, \etc. The bi-temporal images in CLCD were collected by Gaofen-2 in Guangdong Province, China, in 2017 and 2019, respectively, with spatial resolutions ranging from 0.5 to 2 m. Following the standard procedure detailed in \cite{liu2023attention}, each image in the dataset is segmented into 256$\times$256 patches. Consequently, the CLCD dataset is divided into 1440, 480, and 480 pairs for training, validation, and testing, respectively.

\subsubsection{OSCD}
\label{oscd}
The OSCD dataset \cite{daudt2018urban_OSCD} is a relatively low-resolution dataset with a resolution ranging from 10m to 60m. It was captured by the Sentinel-2 satellites in various countries with different levels of urbanization and has experienced urban growth or changes. This resolution enables the detection of large buildings in the image pairs. However, smaller changes such as the appearance of small buildings, extensions of existing buildings, or additions of lanes to roads may not be obvious, making diverse change detection challenging. The dataset consists of 24 regions of approximately 600$\times$600 pixels. In accordance with common practice, each image in the dataset is cropped into 256$\times$256 patches. As a result, the OSCD dataset is divided into 75 training pairs and 28 test pairs.

\subsection{Evaluation Metrics and Compared Methods}
\label{evaluation_metrics_and_compared_methods}

\subsubsection{Evaluation Metrics}
\label{evaluation_metrics}
Following the widely used evaluation protocols in the change detection task, we use three accuracy metrics, \ie, F1 score (\textit{F}1), intersection over union (\textit{IoU}) and overall accuracy (\textit{OA}), to evaluate our proposed method. They as formulated as follows:
\begin{gather}
\label{metrics}
    P = \frac{TP}{TP + FP}, \nonumber \\
    R = \frac{TP}{TP + FN}, \nonumber \\
    F1 = \frac{2PR}{P + R}, \nonumber \\
    IoU = \frac{TP}{TP + FN + FP}, \nonumber \\
    OA = \frac{TP + TN}{TP + TN + FN + FP}.
\end{gather}
where TP, FP, TN, and FN indicate true positive, false positive, true negative, and false negative, respectively. For all the metrics, a higher value means better detection performance.

\subsubsection{Compared Methods}
\label{compared_methods}
To verify the effectiveness of the proposed method, nine representative and open-source methods are selected for comparison which are described as follows:




a) DTCDSCN \cite{liu2020building}: A dual task-constrained deep siamese convolutional network is introduced which can accomplish change detection and semantic segmentation. It applies channel and spatial attention to improve the interactive feature representation.

b) SNUNet \cite{fang2021snunet}: The bi-temporal differential features are extracted by the densely connected siamese network which focuses not only on high-level semantic features but also on the low-level fine-grained features.

c) ChangeFormer \cite{bandara2022transformer}: Multi-scale long-range features are extracted by a hierarchical swin-transformer encoder and decoder with a multi-layer perception.

d) BIT \cite{chen2021remote}: The bi-temporal images are represented as semantic tokens, then using a transformer encoder to model contexts and a transformer decoder to refine the context-rich tokens.

e) ICIFNet \cite{feng2022icif}: An intra-scale cross-interaction and inter-scale feature fusion network that jointly captures spatio-temporal contextual information and obtains short-long range representations of bi-temporal features.

f) DMINet \cite{feng2023change}: An inter-temporal joint-attention module which consists of self-attention and cross-attention block, aims to model the global relations of input images.

g) GASNet \cite{zhang2023global}: This is a CNN-transformer model that uses CNN as the backbone to extract multi-scale features and employs transformer encoder-decoder to model contextual information.

h) AMTNet \cite{liu2023attention}: A global-aware network that models relations between scene and foreground, is proposed to solve the class imbalance problem of change detection task.

i) EATDer \cite{ma2024eatder}: An edge-assisted detector incorporates an edge-aware decoder to integrate the edge information obtained by the encoder, thereby enhancing the feature representation of changed regions.

\begin{table*}[t]
\centering
\caption{Performance comparison of different change detection methods on LEVIR-CD, WHU-CD, and CLCD datasets, respectively. The best results are highlighted in \textbf{bold} and the second best results are \underline{underlined}. All results of the three evaluation metrics are described as percentages (\%).}
\begin{tabular}{c|c|c|ccc|ccc|ccc}
\hline \toprule[0.6pt]
\multirow{2}{*}{Method} & \multirow{2}{*}{\#Params(M)} & \multirow{2}{*}{FLOPs(G)} & \multicolumn{3}{c|}{LEVIR-CD} & \multicolumn{3}{c|}{WHU-CD} & \multicolumn{3}{c}{CLCD} \\
\cline{4-12}
& & & F1 & IoU & OA & F1 & IoU & OA & F1 & IoU & OA \\
\hline
DTCDSCN \cite{liu2020building} & 41.07 & 20.44 & 87.43 & 77.67 & 98.75 & 79.92 & 66.56 & 98.05 & 57.47 & 40.81 & 94.59 \\
SNUNet \cite{fang2021snunet} & 12.04 & 54.82 & 88.16 & 78.83 & 98.82 & 83.22 & 71.26 & 98.44 & 60.82 & 43.63 & 94.90 \\
ChangeFormer \cite{bandara2022transformer} & 41.03 & 202.79 & 90.40 & 82.48 & 99.04 & 87.39 & 77.61 & 99.11 & 61.31 & 44.29 & 94.98 \\
BIT \cite{chen2021remote} & \textbf{3.55} & \textbf{10.63} & 89.31 & 80.68 & 98.92 & 83.98 & 72.39 & 98.52 & 59.93 & 42.12 & 94.77 \\
ICIFNet \cite{feng2022icif} & 23.82 & 25.36 & 89.96 & 81.75 & 98.99 & 88.32 & 79.24 & 98.96 & 68.66 & 52.27 & 95.77 \\
DMINet \cite{feng2023change} & \underline{6.24} & \underline{14.42} & 90.71 & 82.99 & 99.07 & 88.69 & 79.68 & 98.97 & 67.24 & 50.65 & 95.21 \\
GASNet \cite{zhang2023global} & 23.59 & 23.52 & 90.52 & 83.48 & 99.07 & 91.75 & 84.76 & 99.34 & 63.84 & 46.89 & 94.01 \\
AMTNet \cite{liu2023attention} & 24.67 & 21.56 & 90.76 & 83.08 & 98.96 & 92.27 & 85.64 & 99.32 & 75.10 & 60.13 & 96.45 \\
EATDer \cite{ma2024eatder} & 6.61 & 23.43 & 91.20 & 83.80 & 98.75 & 90.01 & 81.97 & 98.58 & 72.01 & 56.19 & 96.11 \\
\hline
\textbf{ChangeViT-T} & 11.68 & 27.15 & \underline{91.81} & \underline{84.86} & \underline{99.17} & \underline{94.53} & \underline{89.63} & \underline{99.57} & \underline{77.31} & \underline{63.01} & \underline{96.67} \\
\textbf{ChangeViT-S} & 32.13 & 38.80 & \textbf{91.98} & \textbf{85.16} & \textbf{99.19} & \textbf{94.84} & \textbf{90.18} & \textbf{99.59} & \textbf{77.57} & \textbf{63.36} & \textbf{96.79} \\
\hline \toprule[0.6pt]
\end{tabular}
\label{tab:overall_results}
\end{table*}

\begin{table}[t]
\centering
\caption{Performance comparison of different change detection methods on the OSCD dataset. The best results are highlighted in \textbf{bold} and the second best results are \underline{underlined}. All results of the three evaluation metrics are described as percentages (\%).}
\begin{tabular}{c|ccc}
\hline \toprule[0.6pt]
\multirow{2}{*}{Method} & \multicolumn{3}{c}{OSCD} \\
\cline{2-4}
& F1 & IoU & OA \\
\hline
DTCDSCN \cite{liu2020building} & 36.13 & 22.05 & 94.50 \\
SNUNet \cite{fang2021snunet} & 27.02 & 15.62 & 93.81 \\
ChangeFormer \cite{bandara2022transformer} & 38.22 & 23.62 & 94.53 \\
BIT \cite{chen2021remote} & 29.58 & 17.36 & 90.15 \\
ICIFNet \cite{feng2022icif} & 23.03 & 13.02 & 94.61 \\
DMINet \cite{feng2023change} & 42.23 & 26.76 & 95.00 \\
GASNet \cite{zhang2023global} & 10.71 & 5.66 & 91.52 \\
AMTNet \cite{liu2023attention} & 10.25 & 5.40 & 94.29 \\
EATDer \cite{ma2024eatder} & 54.23 & 36.98 & 93.85 \\
\hline
\textbf{ChangeViT-T} & \underline{55.13} & \underline{38.06} & \underline{95.01} \\
\textbf{ChangeViT-S} & \textbf{55.51} & \textbf{38.42} & \textbf{95.05} \\
\hline \toprule[0.6pt]
\end{tabular}
\label{tab:overall_results_low_resolution}
\end{table}

\subsection{Comparison with State-of-the-Art Approaches}
\label{subsec:overall_results}
As illustrated in Tab.~\ref{tab:overall_results}, we compare ChangeViT with the previous methods on three high-resolution datasets, \ie, LEVIR-CD, WHU-CD, and CLCD. Notably, all the compared methods employed hierarchical backbone as primary feature extractors. Specifically, DTCDSCN, BIT, ICIFNet, DMINet, GASNet and AMTNet apply ResNet \cite{he2016deep} or its variants \cite{hu2018squeeze_senet} as backbones, SNUNet and EATDer apply nested UNet \cite{ronneberger2015u_unet} and stack non-local blocks \cite{wang2018non_nonlocal}, respectively. In contrast, our approach employs a non-hierarchical, plain ViT as the core backbone which includes ViT-T and ViT-S, coupling with a lightweight detail-capture module which serves as an auxiliary network. From Tab.~\ref{tab:overall_results}, we can summarize the following valuable findings: (1) ChangeViT consistently outperforms the existing works across all datasets and evaluation metrics, despite utilizing the tiny backbone of ViT, which demonstrates its effectiveness. (2) The primary feature extractor in ViTs, despite being non-hierarchical, demonstrates competitive performance when compared to hierarchical-based methods. This underscores the robust feature extraction and representation capabilities that large-scale pre-training ViT can offer, fully realizing its potential. (3) Notably, ChangeViT-T and ChangeViT-S exhibit significant performance gains over the SOTA method (\ie, AMTNet) by 3.99\% and 4.54\% IoU on the WHU-CD dataset. This finding is sensible given that the changes in WHU-CD vary widely, with fewer medium-sized objects compared to smaller and larger ones. This observation aligns with the results illustrated in the middle of Fig.~\ref{subfig:intro_cds_v2c}, underscoring the efficacy of our proposed method in capturing global features and extracting fine-grained spatial information. (4) With an increase in the size of the primary feature extractor, ChangeViT demonstrates enhanced performance. Notably, the detail-capture module, comprising just 2.7M parameters, stands out for its lightweight nature when compared to the total parameter count of each model (\ie, 11.68M and 32.13M). Our proposed ChangeViT achieves a superior balance between efficiency and effectiveness compared to previous methods, underscoring its superiority.

As shown in Tab.~\ref{tab:overall_results_low_resolution}, we also compare ChangeViT with several existing methods on the low-resolution dataset, \ie, OSCD. The targets in the OSCD are relatively smaller than those in high-resolution datasets, exacerbating the foreground-background imbalance issue and making it challenging to detect smaller targets. From Tab.~\ref{tab:overall_results_low_resolution}, the following key points can be noted: (1) The proposed ChangeViT outperforms all compared methods across three evaluation metrics, despite utilizing tiny or small models of ViT, demonstrating its effectiveness on the low-resolution dataset. (2) GASNet and AMTNet perform poorly on this dataset, likely due to their inefficiency in detecting small targets. Although GASNet introduces a foreground-awareness module to address the category imbalance between the foreground and background, it still underperforms in detecting changes in low-resolution remote sensing images.

\subsection{Diagnostic Study}
\label{subsec:ablation_experiments}

\begin{table*}[ht]
\centering
\caption{Study the effectiveness of proposed modules with different transformer architectures on three datasets, respectively. The check mark (\checkmark) denotes combining with our proposed modules. All results are described as percentages (\%).}
\begin{tabular}{c|c|ccc|ccc|ccc}
\hline \toprule[0.6pt]
\multirow{2}{*}{Backbone} & \multirow{2}{*}{Ours} & \multicolumn{3}{c|}{LEVIR-CD} & \multicolumn{3}{c|}{WHU-CD} & \multicolumn{3}{c}{CLCD} \\
\cline{3-11}
& & F1 & IoU & OA & F1 & IoU & OA & F1 & IoU & OA \\
\hline
\multirow{2}{*}{Swin-S} &  & 89.40 & 80.83 & 98.94  & 93.03 & 86.98 & 99.22 & 73.80 & 58.47 & 96.33 \\
& \checkmark & \textbf{90.18} & \textbf{82.11} & \textbf{99.01} & \textbf{94.04} & \textbf{88.75} & \textbf{99.30} & \textbf{75.41} & \textbf{60.52} & \textbf{96.40} \\
\hline
\multirow{2}{*}{PVT-S} & & 84.60 & 73.31 & 98.38 & 87.36 & 77.55 & 98.89 & 70.25 & 54.15 & 95.76 \\
& \checkmark & \textbf{87.26} & \textbf{77.40} & \textbf{98.68} & \textbf{89.09} & \textbf{80.32} & \textbf{98.92} & \textbf{71.95} & \textbf{56.19} & \textbf{95.90} \\
\hline
\multirow{2}{*}{PiT-S} & & 84.94 & 73.83 & 98.38 & 87.34 & 77.53 & 98.89 & 70.01 & 53.86 & 95.88 \\
& \checkmark & \textbf{87.20} & \textbf{77.31} & \textbf{98.63} & \textbf{89.50} & \textbf{81.00} & \textbf{98.94} & \textbf{72.80} & \textbf{57.23} & \textbf{95.93} \\
\hline
\multirow{2}{*}{ViT-S} & & 82.39 & 70.05 & 98.25 & 84.70 & 73.46 & 98.82 & 69.05 & 52.74 & 95.75 \\
& \checkmark & \textbf{91.98} & \textbf{85.16} & \textbf{99.19} & \textbf{94.84} & \textbf{90.18} & \textbf{99.59} & \textbf{77.53} & \textbf{63.30} & \textbf{96.76} \\
\hline \toprule[0.6pt]
\end{tabular}
\label{tab:influence_of_different_architectures}
\end{table*}

\begin{table*}[ht]
\centering
\caption{Study the effectiveness of proposed modules in ChangeViT on three datasets, respectively. DC and FI denote the detail-capture module and feature injector, respectively. All results are described as percentages (\%).}
\begin{tabular}{ccc|ccc|ccc|ccc}
\hline \toprule[0.6pt]
\multicolumn{3}{c|}{Model} & \multicolumn{3}{c|}{LEVIR-CD} & \multicolumn{3}{c|}{WHU-CD} & \multicolumn{3}{c}{CLCD} \\
\hline
ViT & DC & FI & F1 & IoU & OA & F1 & IoU & OA & F1 & IoU & OA \\
\hline
\checkmark & & & 82.39 & 70.05 & 98.25 & 84.70 & 73.46 & 98.82 & 69.18 & 52.88 & 95.68 \\
& \checkmark & & 88.12 & 78.76 & 98.80 & 90.20 & 82.15 & 99.24 & 69.72 & 53.51 & 95.98 \\
\checkmark & \checkmark & & 91.20 & 83.82 & 99.11 & 93.30 & 87.43 & 99.46 & 75.36 & 60.46 & 96.62 \\
\checkmark & \checkmark & \checkmark & \textbf{91.98} & \textbf{85.16} & \textbf{99.19} & \textbf{94.84} & \textbf{90.18} & \textbf{99.59} & \textbf{77.53} & \textbf{63.30} & \textbf{96.76} \\

\hline \toprule[0.6pt]
\end{tabular}
\label{tab:effectiveness_of_modules}
\end{table*}

\begin{table*}[ht]
\centering
\caption{Investigation on the impact of multiple scales in the detail-capture module on three datasets, respectively. All results are described as percentages (\%).}
\begin{tabular}{ccc|ccc|ccc|ccc}
\hline \toprule[0.6pt]
\multicolumn{3}{c|}{Scales} & \multicolumn{3}{c|}{LEVIR-CD} & \multicolumn{3}{c|}{WHU-CD} & \multicolumn{3}{c}{CLCD} \\
\hline
1/8 & 1/4 & 1/2 & F1 & IoU & OA & F1 & IoU & OA & F1 & IoU & OA \\
\hline
\checkmark & & & 91.32 & 84.03 & 99.12 & 94.20 & 89.04 & 99.55 & 76.43 & 61.85 & 96.49 \\
& \checkmark & & 91.08 & 83.62 & 99.10 & 92.90 & 86.74 & 99.45 & 73.25 & 57.79 & 96.31 \\
& & \checkmark & 89.43 & 80.89 & 98.94 & 90.87 & 83.28 & 99.29 & 70.82 & 54.82 & 95.79 \\
\checkmark & \checkmark & & 91.56 & 84.43 & 99.15 & 94.25 & 89.07 & 99.58 & 77.30 & 63.07 & 96.62 \\
\checkmark & & \checkmark & 91.45 & 94.24 & 99.14 & 94.02 & 88.67 & 99.44 & 77.10 & 62.73 & 96.56 \\
& \checkmark & \checkmark & 90.94 & 83.39 & 99.09 & 92.49 & 86.04 & 99.41 & 75.22 & 60.28 & 96.46 \\
\checkmark & \checkmark & \checkmark & \textbf{91.98} & \textbf{85.16} & \textbf{99.19} & \textbf{94.84} & \textbf{90.18} & \textbf{99.59} & \textbf{77.53} & \textbf{63.30} & \textbf{96.76} \\

\hline \toprule[0.6pt]
\end{tabular}
\label{tab:influence_of_detail_capture}
\end{table*}

\textbf{Effectiveness with different architectures.}
\label{influence_of_different_architectures}
In Tab.~\ref{tab:influence_of_different_architectures}, we investigate the effectiveness of the proposed modules with different architectures, including hierarchical (\ie, Swin-S \cite{liu2021swin}, PVT-S \cite{wang2021pyramid_pvt}, PiT-S \cite{heo2021rethinking_pit}) and non-hierarchical (\ie, ViT-S \cite{dosovitskiy2020image}) transformers. Key observations from the table include: (1) Without combining with our proposed modules, the non-hierarchical ViT-S underperformers the other hierarchical methods across all metrics on three datasets. (2) When integrated with our proposed modules, all transformers exhibit performance improvements, indicating the efficacy of our approach regardless of transformer architectures. (3) ViT-S achieves significant performance gains over hierarchical transformers when equipped with our proposed modules, suggesting that our modules effectively mitigate ViT-S's limitations in capturing detailed information to detect smaller objects.

\textbf{Effectiveness of proposed modules.}
\label{effectiveness_of_proposed_modules}
To investigate the effectiveness of the proposed modules, we conduct comprehensive diagnostic experiments on three datasets. As shown in Tab.~\ref{tab:effectiveness_of_modules}, we take various combinations of components into account and explore the contribution of each module. We apply ViT as baseline, which consists of a plain ViT and a decoder. Coupling with detail-capture module, ViT can unleash its potential and improve 8.81\%, 8.60\%, and 6.18\% F1 on three datasets compared to the baseline, which indicates that the detail-capture module can supplement the detailed spatial information, which is essential for the change detection task. Furthermore, when combined with the feature injector, there are additional performance gains of 0.78\%, 1.54\%, and 2.17\% in F1,  indicating the effectiveness of incorporating detailed information at higher levels. In summary, all of our proposed modules are essential and effective in the ChangeViT framework.

\textbf{Impact of multiple scales.}
\label{impact_of_multiple_scales}
To investigate the necessity of capturing multiple scales in the detail-capture module, we conduct experiments using multi-scale features, \ie, 1/2, 1/4, 1/8. As shown in Tab.~\ref{tab:influence_of_detail_capture}, we can get the following key observations: (1) Single-scale features often yield subpar results, while the amalgamation of multi-scale features leads to enhanced performance. (2) An interesting finding is that high-level features or their combinations can achieve better performance than low-level features. 
(3) Furthermore, the inclusion of three-scale features results in mutual improvements, indicating that multi-scale features leverage spatial cues across complementary levels.

\textbf{Impact of pre-trained weights.}
\label{influence_of_pretrained_weights}
To investigate the impact of pre-trained weights on ChangeViT, we apply various model initialization approaches, including random initialization and several publicly available large-scale pre-trained weights derived from both supervised and self-supervised training strategies on various datasets. As illustrated in Tab.~\ref{tab:influence_of_pretrained_weights}, we observed the following key points: (1) Both ChangeViT-T and ChangeViT-S exhibit improved detection accuracy when utilizing pre-trained weights compared to random initialization. (2) DINOv2-S provides the most effective pre-trained weights for the ChangeViT-S model, benefiting from large-scale data pre-training. (3) When DMINet, GASNet, AMTNet, and ChangeViT are pre-trained on the same data, \ie, ImageNet-1k, the proposed ChangeViT outperforms all the CNN-based methods, demonstrating the effectiveness of transferring the priorities of large pre-trained ViT models to the change detection task.

\begin{table*}[tb]
\setlength\tabcolsep{3pt}
\centering
\caption{Study the impact of different pre-trained weights of ViT on three datasets, respectively. All results are described as percentages (\%).}
\begin{tabular}{c|c|c|c|c|ccc|ccc|ccc}
\hline \toprule[0.6pt]
\multirow{2}{*}{Model} & \multirow{2}{*}{Backbone} & \multirow{2}{*}{Pretrain} & \multirow{2}{*}{Pre-trained Data} & Training & \multicolumn{3}{c|}{LEVIR-CD} & \multicolumn{3}{c|}{WHU-CD} & \multicolumn{3}{c}{CLCD} \\
\cline{6-14}
& & & & Strategy & F1 & IoU & OA & F1 & IoU & OA & F1 & IoU & OA \\
\hline
DMINet \cite{feng2023change} & ResNet18 & - & ImageNet(1k) & Supervised & 90.71 & 82.99 & 99.07 & 88.69 & 79.68 & 98.97 & 67.24 & 50.65 & 95.21 \\
GASNet \cite{zhang2023global} & ResNet34 & - & ImageNet(1k) & Supervised & 90.52 & 83.48 & 99.07 & 91.75 & 84.76 & 99.34 & 63.84 & 46.89 & 94.01 \\
AMTNet \cite{liu2023attention} & ResNet50 & - & ImageNet(1k) & Supervised & 90.76 & 83.08 & 98.96 & 92.27 & 85.64 & 99.32 & 75.10 & 60.13 & 96.45 \\
\hline
\hline
\multirow{2}{*}{ChangeViT-T} & \multirow{2}{*}{ViT(Tiny)} & Random Init & - & - & 91.58 & 84.47 & 99.15 & 93.78 & 88.29 & 99.51 & 76.91 & 62.49 & 96.66 \\
& & DeiT-T \cite{touvron2021training_deit} & ImageNet(1k) & Supervised & \textbf{91.81} & \textbf{84.86} & \textbf{99.17} & \textbf{94.53} & \textbf{89.63} & \textbf{99.57} & \textbf{77.31} & \textbf{63.01} & \textbf{96.67} \\
\hline
\multirow{5}{*}{ChangeViT-S} & \multirow{5}{*}{ViT(Small)} & Random Init & - & - & 90.82 & 83.19 & 99.09 & 93.65 & 88.06 & 99.50 & 75.05 & 60.06 & 96.59 \\
& & DeiT-S \cite{touvron2021training_deit} & ImageNet(1k) & Supervised & 91.78 & 84.81 & 99.17 & 94.73 & 89.99 & 99.58 & 77.24 & 62.69 & 96.68 \\
& & DINO-S \cite{dino} & ImageNet(w/o labels) & Self-supervised & 91.68 & 84.64 & 99.16 & 94.70 & 89.94 & 99.58 & 77.05 & 62.67 & 96.65 \\
& & \multirow{2}{*}{DINOv2-S \cite{oquab2023dinov2}} & ImageNet(1k, 22k) \& & \multirow{2}{*}{Self-supervised} & \multirow{2}{*}{\textbf{91.98}} & \multirow{2}{*}{\textbf{85.16}} & \multirow{2}{*}{\textbf{99.19}} & \multirow{2}{*}{\textbf{94.84}} & \multirow{2}{*}{\textbf{90.18}} & \multirow{2}{*}{\textbf{99.59}} & \multirow{2}{*}{\textbf{77.53}} & \multirow{2}{*}{\textbf{63.30}} & \multirow{2}{*}{\textbf{96.76}} \\
& & & Google Landmarks & & & & & & & & & & \\

\hline \toprule[0.6pt]
\end{tabular}
\label{tab:influence_of_pretrained_weights}
\end{table*}

\begin{figure*}[htb]
\centering
\includegraphics[width=0.9\linewidth]{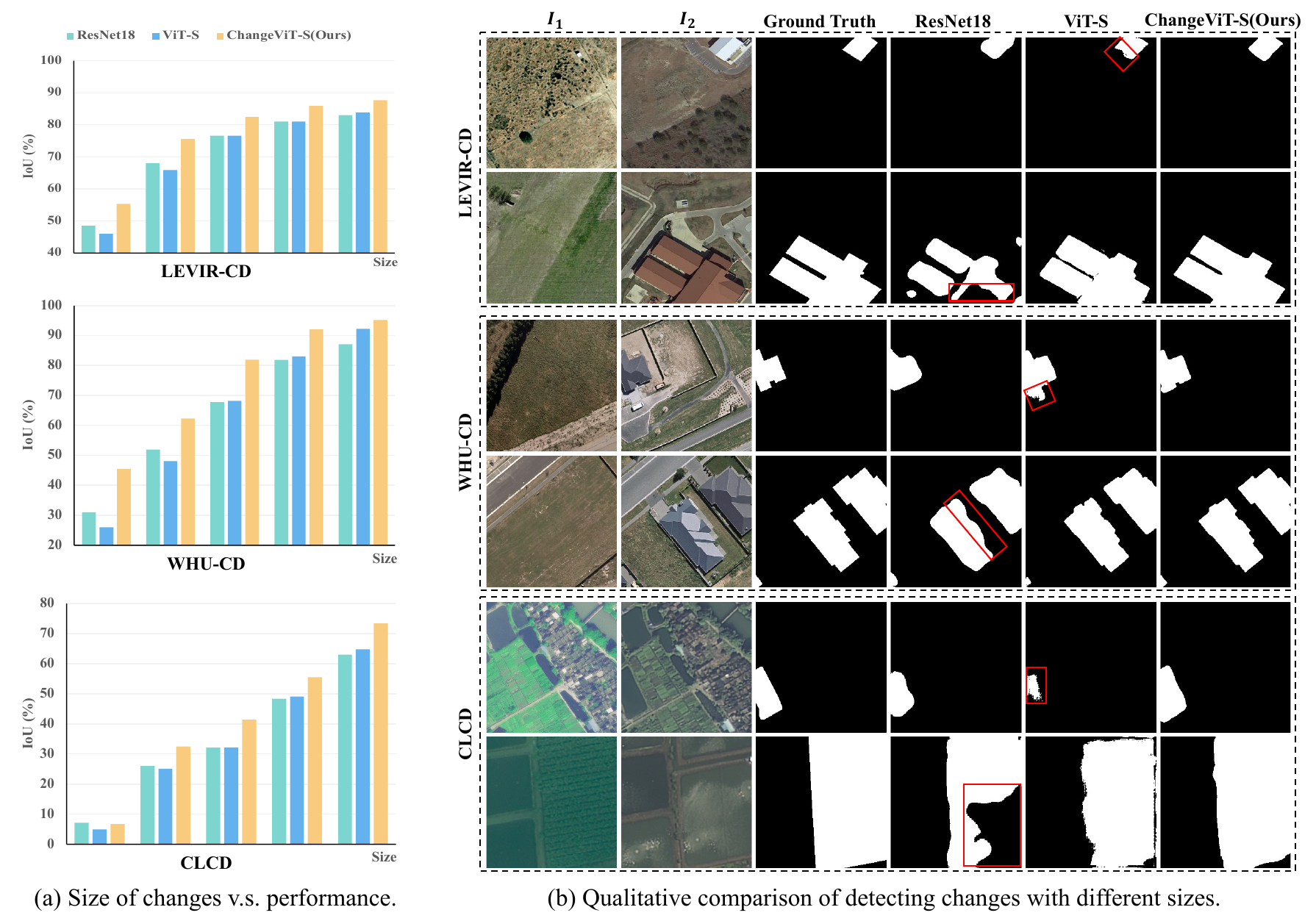}
\caption{(a) Each dataset is split into five intervals on average based on the change sizes. The horizontal axis incrementally reflects the change sizes, progressing from smaller to larger changes. (b) The predicted map within the red box indicates a poor detection outcome.
}
\label{fig:quantity_results}
\end{figure*}

\textbf{Choice of query, key and value.}
\label{{choice_of_query_key_and_value}}
Two experiments are conducted to investigate different modeling approaches in the feature injector, as shown in Tab.~\ref{tab:influence_of_structure_enhancement}. In the first experiment, $F_{V}$ serves as query, and $F_{C}$ serves as key and value, yielding the best performance. This result is consistent with the conjecture mentioned in Sec.~\ref{subsec:structure_enhancement}, suggesting that the feature injector effectively captures low-level value information most relevant to the high-level query and reintegrates it back to the query. Therefore, through cross-attention, high-level fine-grained features can seamlessly merge with low-level features.

\begin{table*}[htb]
\centering
\caption{Study the impact of different modeling approaches in the feature injector on three datasets, respectively. All results are described as percentages (\%).}
\begin{tabular}{c|c|ccc|ccc|ccc}
\hline \toprule[0.6pt]
\multirow{2}{*}{Query} & \multirow{2}{*}{Key\&Value} & \multicolumn{3}{c|}{LEVIR-CD} & \multicolumn{3}{c|}{WHU-CD} & \multicolumn{3}{c}{CLCD} \\
\cline{3-11}
& & F1 & IoU & OA & F1 & IoU & OA & F1 & IoU & OA \\
\hline
$F_{V}$ & $F_{C}$ & \textbf{91.98} & \textbf{85.16} & \textbf{99.19} & \textbf{94.84} & \textbf{90.18} & \textbf{99.59} & \textbf{77.53} & \textbf{63.30} & \textbf{96.76} \\
$F_{C}$ & $F_{V}$ & 91.78 & 84.80 & 99.17 & 94.60 & 89.75 & 99.58 & 75.84 & 61.08 & 96.58 \\
\hline \toprule[0.6pt]
\end{tabular}
\label{tab:influence_of_structure_enhancement}
\end{table*}

\textbf{Size of changes v.s. performance.}
As depicted in Fig.~\ref{fig:quantity_results}~(a), we conduct experiments on three datasets using the detail-capture module, ViT-S, and our proposed method to quantitatively analyse the performance of each method under different change sizes. The detail-capture module and ViT-S both integrate with a decoder which is the same as ChangeViT. The results indicate that the detail-capture module excels at detecting smaller changed targets, while the ViT-S demonstrates superiority in detecting larger ones. Our method capitalizes on ViT's powerful feature expression while leveraging a detail-capture module for fine-detail information mining. This comprehensive approach enables superior performance across targets of all sizes.

\textbf{Qualitative results.}
We present representative visualization results on three datasets, comparing the performance of the detail-capture module, ViT-S, and our proposed method to demonstrate the effectiveness of ChangeViT. As shown in Fig.~\ref{fig:quantity_results}~(b), the first row in each dataset presents the test results for smaller targets, while the second row corresponds to larger targets. From the results, we can see that the detail-capture module excels at detecting smaller targets, whereas ViT-S demonstrates superiority in detecting larger targets. The fundamental distinction lies in the local receptive field of CNN, enabling them to extract intricate local features, while ViT possesses a global receptive field, facilitating the extraction of comprehensive global information. The proposed method efficiently integrates global and local information, resulting in superior performance. 

\begin{figure*}[ht]
\centering
\includegraphics[width=0.9\linewidth]{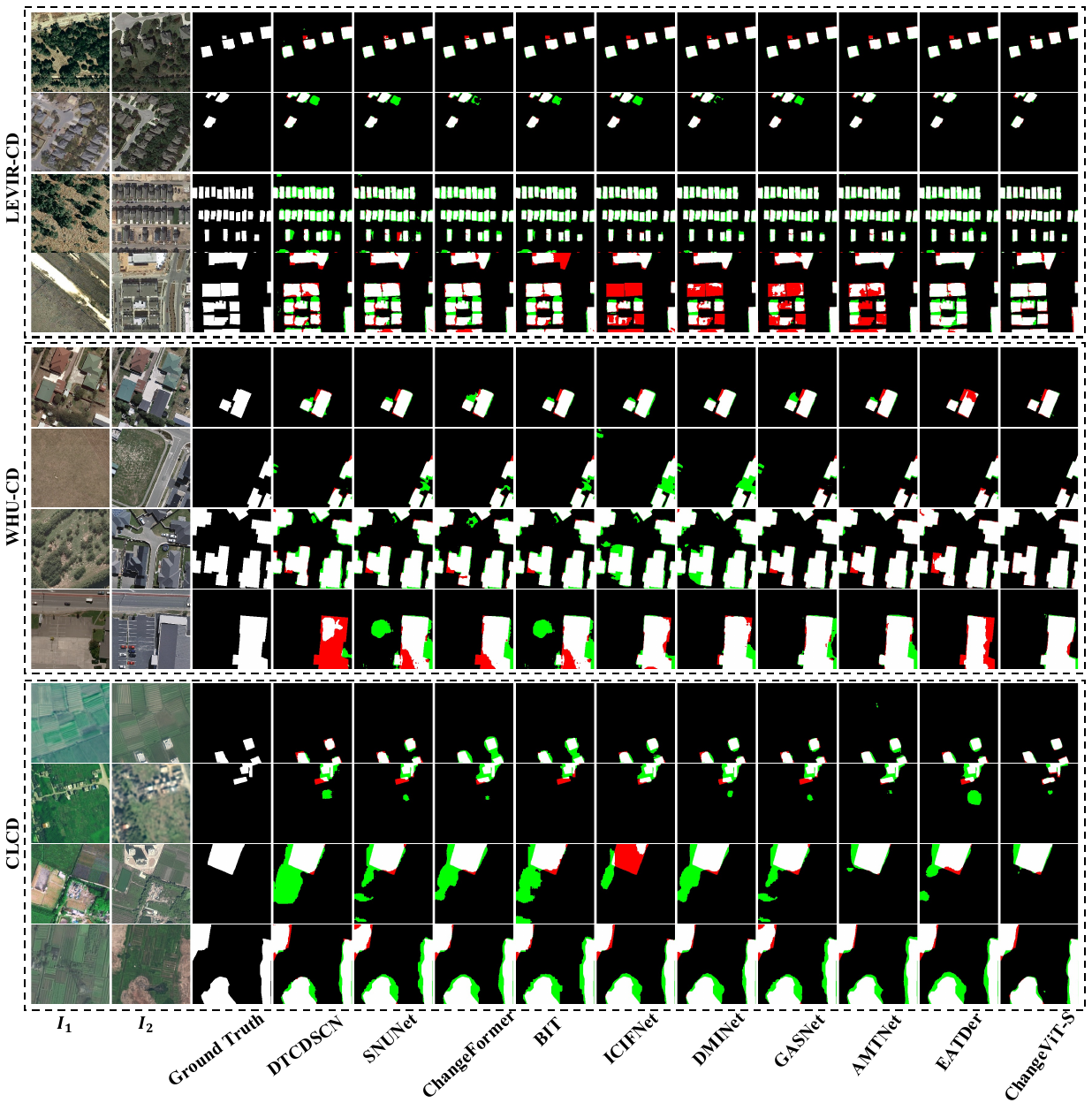}
\caption{Qualitative comparison of different methods on the three datasets. White represents a true positive, black is a true negative, \textcolor{green}{green} indicates a false positive, and \textcolor{red}{red} is a false negative. Fewer \textcolor{green}{green} and \textcolor{red}{red} pixels represent better performance. For better clarity, please zoom in on the figure.
}
\label{fig:compared_methods_visualization}
\end{figure*}

To qualitatively compare with previous methods, we provide comprehensive samples encompassing small, large, sparse, and dense targets, as illustrated in Fig.~\ref{fig:compared_methods_visualization}. From these samples, several key observations emerge intuitively: (1) Our proposed method consistently outperforms all compared methods across various change sizes. This is attributed to the robust global modeling capabilities of ViT and the detail-capture module's capacity to extract intricate spatial information. Additionally, a feature injector integrates low-level fine-grained spatial features into ViT's high-level semantic representations, enhancing ChangeViT's capability to detect changes of diverse sizes. (2) In detecting dense objects, regardless of their size, ChangeViT consistently delineates clear boundaries compared to prior methods. This underscores ChangeViT's effectiveness in capturing both global semantic information and local spatial details of neighboring objects.

\section{Conclusion}
\label{sec:conclusion}
In this paper, we present a simple yet effective framework, namely ChangeViT, that leverages the plain ViT as its primary feature extractor to capture large-scale changes. Coupled with a detail-capture module dedicated to fine-grained spatial features, ChangeViT seamlessly integrates these details into ViT's feature representation through the cross-attention mechanism.  Experimental results demonstrate ChangeViT's supremacy over meticulously designed hierarchical models across all evaluation metrics on four widely adopted datasets, highlighting the untapped potential of vanilla ViTs for change detection. Furthermore, comprehensive diagnostic analyses and visualization results provide insights into the contribution of each module. We aim for this study to offer valuable insights to the research community and ignite further exploration into leveraging vanilla ViTs for other related computer vision tasks, such as change caption.

{\small
\bibliographystyle{IEEEtran}
\bibliography{egbib}
}

\begin{IEEEbiography}[{\includegraphics[width=1in,height=1.25in,clip,keepaspectratio]{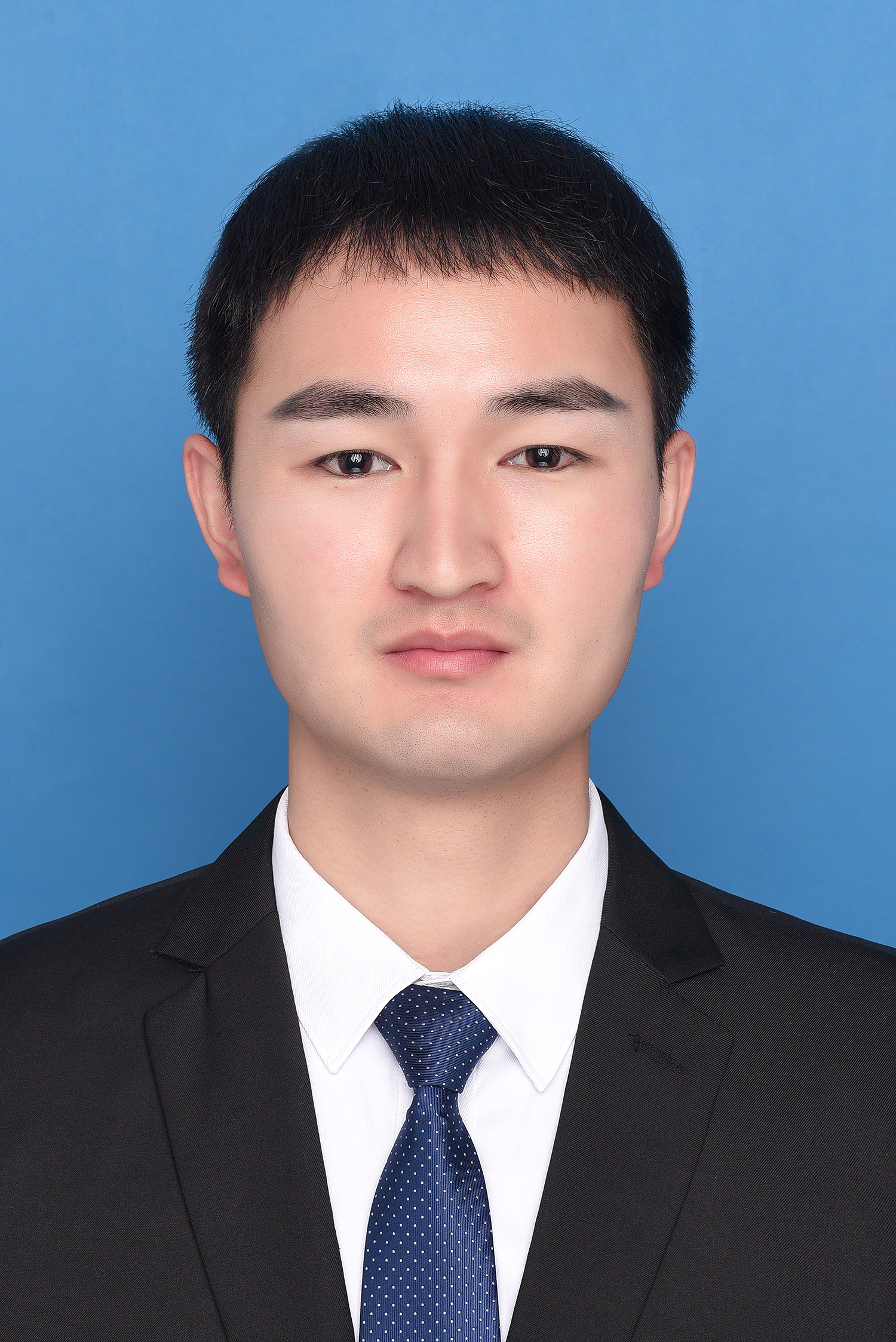}}]{Duowang Zhu} received the B.S. and M.S. degree in School of Electronic Information and Communications from Huazhong University of Science and Technology (HUST), Wuhan, China, in 2020 and 2022, respectively. He is currently pursuing the Ph.D. degree with the State Key Laboratory of Information Engineering in Surveying, Mapping and Remote Sensing (LIESMARS), Wuhan University, Wuhan, China. His current research areas include computer vision and remote sensing image process.
\end{IEEEbiography}

\begin{IEEEbiography}[{\includegraphics[width=1in,height=1.25in,clip,keepaspectratio]{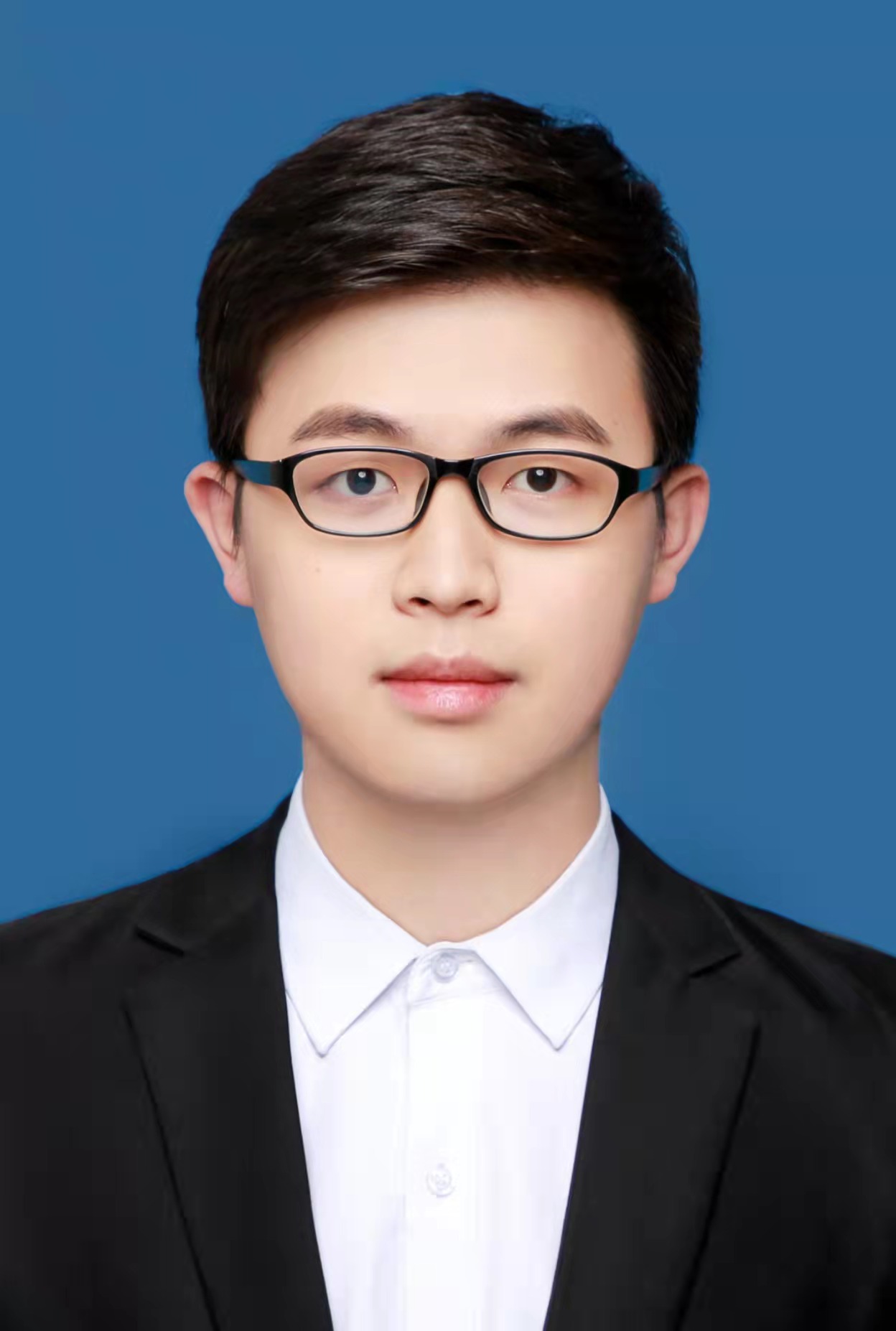}}] {Xiaohu Huang} received the B.S. and M.S. degree in School of Electronic Information and Communications from Huazhong University of Science and Technology (HUST), Wuhan, China, in 2020 and 2023, respectively. Now, he is pursuing the Ph.D degree in the University of Hong Kong (HKU), Hong Kong, China. His current research areas include computer vision and machine learning.
\end{IEEEbiography}

\begin{IEEEbiography}[{\includegraphics[width=1in,height=1.25in,clip,keepaspectratio]{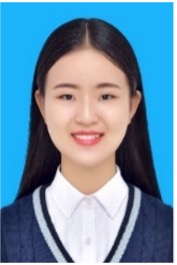}}]{Haiyan Huang} received the M.S. degree Huazhong University of Science and Technology, Wuhan, China, in 2022. She is currently studying for the Ph.D. degree with the State Key Laboratory of Information Engineering in Surveying, Mapping and Remote Sensing, Wuhan University. Her research interests include high spatial resolution remote sensing image understanding and analysis.
\end{IEEEbiography}

\begin{IEEEbiography}
[{\includegraphics[width=1in,height=1.25in,clip,keepaspectratio]{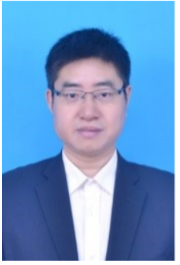}}]{Zhenfeng Shao} received the Ph.D. degree in photogrammetry and remote sensing from Wuhan University, Wuhan, China, in 2004. Since 2009, he has been a Full Professor with the State Key Laboratory of Information Engineering in Surveying, Mapping and Remote Sensing, Wuhan University. He has authored or coauthored more than 50 peer-reviewed articles in international journals. His research interests include high-resolution image processing, pattern recognition, and urban remote sensing applications. Dr. Shao was a recipient of the Talbert Abrams Award for the Best Paper in Image Matching from the American Society for Photogrammetry and Remote Sensing in 2014 and the New Century Excellent Talents in University from the Ministry of Education of China in 2012. Since 2019, he has been serving as an Associate Editor for the Photogrammetric Engineering \& Remote Sensing (PE \& RS) specializing in smart cities, photogrammetry, and change detection.
\end{IEEEbiography}

\begin{IEEEbiography}
[{\includegraphics[width=1in,height=1.25in,clip,keepaspectratio]{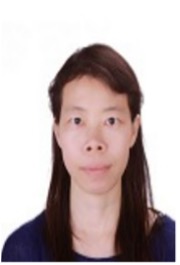}}]{Qimin Cheng} received the Ph.D. degree in cartography and geographic information system from the Institute of Remote Sensing Applications, Chinese Academy of Sciences, Beijing, China, in 2004. She is currently a Professor with the Huazhong University of Science and Technology, Wuhan, China. Her research interests include image retrieval and annotation, and remote sensing images understanding and analysis.
\end{IEEEbiography}

\vfill

\end{document}